\documentclass[journal,twocolumn]{IEEEtran}
\ifCLASSINFOpdf
\else
\fi
\usepackage{lipsum}
\usepackage{xcolor}
\usepackage{graphicx}
\usepackage{amsmath}
\usepackage{siunitx}
\usepackage{algorithm}
\usepackage{algpseudocode}
\usepackage[top=1in, left=0.75in, right=0.75in, bottom=0.75in]{geometry}
\begin{document}
%
\title{Machine Learning-Driven Burrowing with a Snake-Like Robot}
%
%
%

\author{Sean Even*, Holden Gordon*, Hoeseok Yang, Yasemin Ozkan-Aydin \vspace{-3em}
\thanks{* These authors contributed equally.}%
\thanks{S. Even and Y. Ozkan-Aydin are with the University of Notre Dame}
\thanks{H. Gordon and H. Yang are with Santa Clara University}
}

\maketitle
\begin{abstract}
    Subterranean burrowing is inherently difficult for robots because of the high forces experienced as well as the high amount of uncertainty in this domain. Because of the difficulty in modeling forces in granular media, we propose the use of a novel machine-learning control strategy to obtain optimal techniques for vertical self-burrowing. In this paper, we realize a snake-like bio-inspired robot that is equipped with an IMU and two triple-axis magnetometers. Utilizing magnetic field strength as an analog for depth, a novel deep learning architecture was proposed based on sinusoidal and random data in order to obtain a more efficient strategy for vertical self-burrowing. This strategy was able to outperform many other standard burrowing techniques and was able to automatically reach targeted burrowing depths. We hope these results will serve as a proof of concept for how optimization can be used to unlock the secrets of navigating in the subterranean world more efficiently.
\end{abstract}

\section{Introduction}

The task of excavation and subterranean penetration presents substantial challenges owing to the dynamic forces exerted by the subterranean environment \cite{li2013terradynamics,pradhan2019mass,maladen2011mechanical}. This has driven numerous organisms to evolve specialized mechanisms aimed at reducing energy consumption, thereby enabling them to achieve precise outcomes without resorting to extensive soil displacement\cite{martinez2022bio}. For example, plant roots employ strategies such as tip extension and circumnutation to efficiently gather nutrients from beneath the soil \cite{roots}. In the animal kingdom, body structures such as asymmetric head shape and limb movements to create granular fluidization help biological organisms move through the soil more effectively \cite{Naclerio}. Furthermore, soft-bodied organisms such as earthworms utilize gaits such as retrograde peristalsis to achieve motion through soil by reappropriating body mass \cite{omori2008locomotion}.

The creation of burrowing robots is an emerging  field of study that develops systems that can be used for applications like emergency response, soil analysis in agriculture, and exploring other planets. Traversing through subterranean environments is quite difficult as the robot must be able to withstand forces that can be orders of magnitude higher than in air and water \cite{Tao_2021}.

While the study of burrowing robots is relatively new, a few important studies have laid the groundwork for fossorial locomotion. These studies largely fall into two classes: soft-burrowing robots and rigid-body burrowing robots. Soft-burrowing robots are inspired by earthworm locomotion and root growth in plants. Liu et al. designed a soft robot that combines Kirigami skin and radially expanding pneumatic actuators to mimic earthworm anchoring mechanisms \cite{Liu}.  Naclerio et al. developed a steerable vine robot for subterranean locomotion, achieving significantly faster burrowing speeds and obstacle navigation using granular fluidization \cite{Naclerio}.

\begin{figure}[!t]
\centering
\includegraphics[width=\columnwidth]{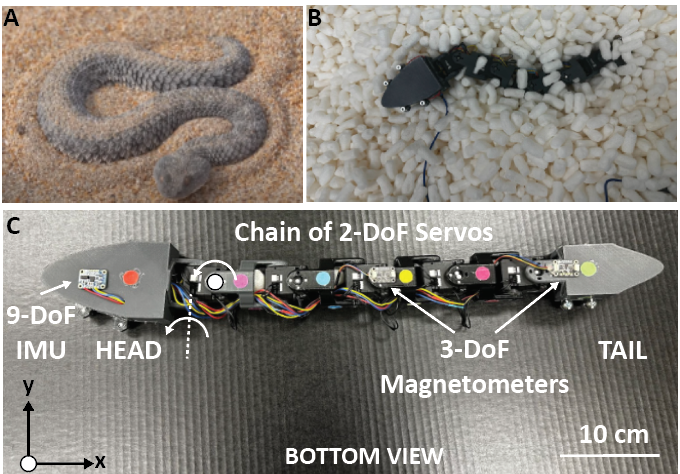}
\caption{{\bf Biological and robotic snake burrowing in granular media. A} Biological example of snake burrowing in the sand, {\bf B} Snake robot in a testing environment filled with packing peanuts, {\bf C} Snake Robot consists of a chain of five Dynamixel 2XL430 servos, two 11.1V batteries located at head and tail, 9-DoF IMU, two 3-DoF magnetometers and an Open-CM controller. The colored stickers show the segments of the robot. The arrows in the first segment of the robot show the direction of the rotation axis of the servo. All the servos at the segments are oriented in the same direction. }
\label{intro_figure}
\end{figure}

Rigid-body burrowing robots have drawn inspiration from animals in nature that dig into the soil such as the mole crab. Russel presented CRABOT, a burrowing robot inspired by mole crabs, designed to locate underground chemical leaks and capable of traversing the surface of granular media \cite{Russell}. Treers et al. presented EMBUR as capable of vertical burrowing in granular media and providing insights into its behavior through parametric studies \cite{Treers}. Bagheri et al. introduced a bio-inspired burrowing robot with various screw and fin designs, highlighting that a one-bladed screw at a lower speed minimizes energy costs for burrowing \cite{Bagheri2023}.

Designing a robot capable of efficient subterranean navigation poses significant challenges, primarily resulting from the complexities associated with modeling the nonlinear and uncertain forces encountered within media \cite{PhysRevE.105.064902}. Traditional control approaches may struggle to accurately model and control such complex systems. However, machine learning techniques can effectively learn from data and adapt to the nonlinearity and uncertainty inherent in the soil environment in real-time, allowing the robot to dynamically adjust its behavior during burrowing.

In this study, we present a snake-like robotic system that consists of a chain of two-degree-of-freedom servo motors, enabling individual sections of the robot to actuate in the pitch and yaw directions with increased flexibility and adaptability during locomotion. The robot's modular design permits the extension or contraction of its body, enabling the system to adapt to various environments and tasks. The snake robot is equipped with two types of sensors. First, the robot’s head is equipped with an internal measurement unit (IMU) to accurately estimate the posture of the robot's head and the burrowing depth of the robot using magnetic field strength. Additionally, the robot is equipped with two triple-axis magnetometers that were added to the center and tail of the bottom face of the robot. Similar to the IMU, data from the magnetometers is used to track the height of the robot as it burrows. The control system of the snake-like robot is implemented on a Jetson Nano, a compact and powerful edge computing platform that hosts an efficient Neural Network, which has been trained to generate optimized instructions for the robot's burrowing.

The primary goal of this study is to design and train a robotic snake capable of self-vertical burrowing within a controlled laboratory, simulating a granular medium using packing peanuts of varying sizes. The robot's task involves starting from the medium's surface and using controlled movements to autonomously submerge itself beneath the granular medium. The objective is to optimize efficiency and minimize the time required for burrowing while maximizing the depth attained.

\begin{figure}[!t]
\centering
\includegraphics[width=\columnwidth]{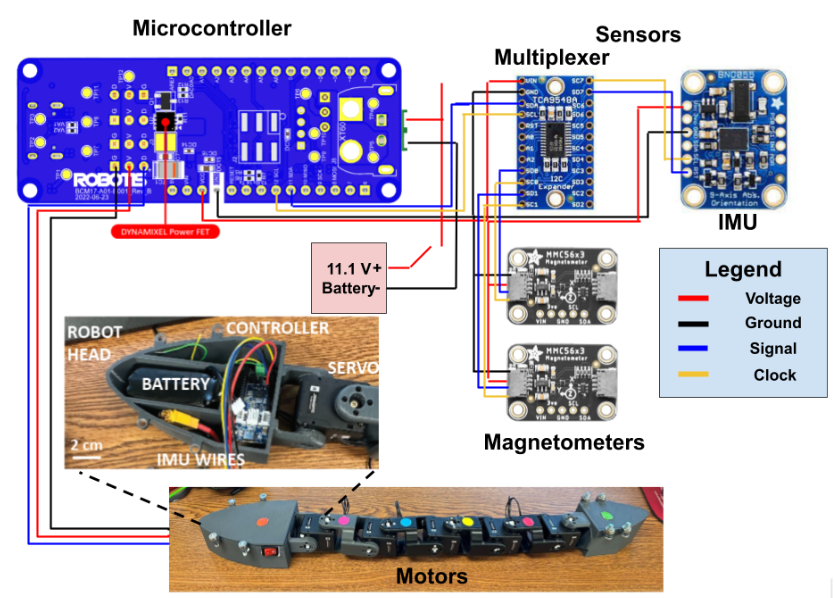}
\caption{\textbf{Electronic schematic of snake robot.} The robot is controlled by a Robotis OpenRB-150 Board. It contains a chain of five  Dynamixel two DoF 2XL430 servos powered by two 11.1 V LiPo batteries located at the head and tail. The robot also contains two Adafruit MMC5603 Triple Axis Magnetometers,a Adafruit BNO055 IMU  and a Adafruit TCA9548A I2C Multiplexer.}
\vspace{-5mm}
\label{schematic}
\end{figure}

\section{Robot Design and Experimental Setup}

The snake robot used in this study consists of a chain of five Dynamixel two DoF 2XL430 servos. These servos give the robot the ability to actuate the pitch and yaw of the robot at segments independently. While these motors only have the capability to actuate the motor about two axes, the combined effect of all these motors generates complex shapes in the three-dimensional space. 

The motors are connected together using 3-D printed parts made from ABS plastic and printed on a Stratasys F170 system. These body connector pieces are designed to connect the yaw motor of one axis to the pitch motor of the subsequent axis. The body of the snake robot was assembled by connecting the five 2XL430 servos together using four such connector pieces.

The motors were controlled using a Robotis Open-RB 150 board. This board is a specialized version of an Arduino MKR Zero board which is optimized to work with the selected servos. In order to power the system, we used Tattu 11.1V 1500 mAh 120C Drone battery to supply high amounts of current due to the number of motors that needed to be powered.

To facilitate the determination of the robot's head orientation and enable the collection of magnetic field data for the purpose of depth measurement during experiments, we attached Adafruit BNO055 9 Degree of Freedom (DoF) Internal Measurement Unit (IMU) to the head of the robot. This device acts as a gyroscope, an accelerometer, and a triple-axis magnetometer.

To enhance the precision of our self-burrowing strategies, we have expanded our data collection efforts to include information regarding the heights of the robot's middle and tail segments of the robot as well. To achieve this, we have used the Adafruit MMC5603 Triple Axis magnetometer, placed to capture magnetic field strength measurements at the lowermost portion of the robot, including the base of its third link and its tail. This dataset of magnetic field strength readings was instrumental in accurately gauging the robot's depth in relation to the underlying surface within the testing environment. Since the sensors shared the same I2C address, we added the Adafruit TCA9548A I2C Mux to connect each of our three sensors to a separate channel, thereby effectively resolving the communication issue encountered with the sensors.

Finally, with all the components selected, we designed a head and tail for the robot that would contain all the selected components. These parts were similarly designed in Solidworks and printed in ABS plastic using a Stratasys F170 3-D printer. The head and tail of the snake contained connectors which enabled them to connect to the body of the snake that was previously described. A full schematic showing all of the components utilized in the robot and how they are connected can be seen in Fig. \ref{schematic}.

 \begin{figure}[!t]
\centering
\includegraphics[width=\columnwidth]{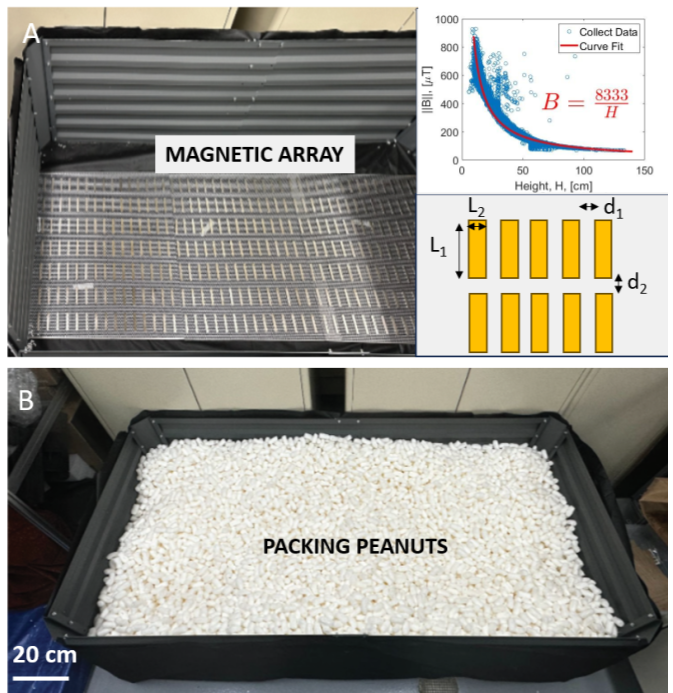}
\caption{\textbf{Experimental setup. A} Empty test environment with an array of Neodymium bar magnets (L$1$=60 mm x L$2$=10 mm x 3 mm) positioned at the bottom with specific horizontal (d$_1$ = 29 mm) and vertical (d$_2$ = 23 mm) spacing.   The relationship between net magnetic field strength $|B|$ obtained by the sensors and the height (H) above the base of the testing environment is given at the top inset. This figure also shows the correspondence between the data collected (blue circles) and the mathematical relationship obtained (red curve).\textbf{B} Tests were conducted in a raised garden bed filled with packing peanuts. }
\vspace{-5mm}
\label{H vs B}
\end{figure}

We tested the robot in an 182.88 cm x 91.44 cm x 60.96 cm (6 ft x 3ft x 2 ft) raised garden bed. At the bottom of this garden bed, we placed an array of evenly spaced 60 mm x 10 mm x 3 mm Neodymium rectangular bar magnets (Fig.\ref{H vs B}A). The magnets were placed in an array such that the polarity of all the bar magnets matched. In this study, magnetic field strength serves as an analog for the depth of the robot at various points. We want to maximize the magnetic field strength measured by the robot's onboard sensors which will correspond to the vertical self-burrowing height of the robot.


Before adding a granular medium to the testing environment, we found a relationship between the magnetic field strength of height above the array of magnets at the base of the enclosure.  We used an apparatus that consisted of an Adafruit MMC603 Triple-axis Magnetometer and an RCWL-1601 Ultrasonic distance sensor. We collected data to determine the correlation between the magnetic field in the z direction of the sensor $B_z$ and the height $H$. Initially, our approach involved utilizing orientation data from the IMU located in the robot's head, combined with known body angles, to exclusively employ the z-component of the magnetic field for height determination. However, when we tried this in practice, the data obtained was not reliable due to the errors in estimating the posture of the robot. So, we decided to use the net magnetic field of the robot, $||B||= \sqrt{B_x^2+B_y^2+B_z^2}$, to characterize the height of the robot. By plotting the data as seen in Fig. \ref{H vs B}A and fitting the curve to it, we obtained $H=\frac{8333}{||B||}$  that relates the height of the sensor to the net magnetic field strength.
\section{Data Collection and Machine Learning}

Robotic applications have become increasingly complex and traditional modeling approaches often fail to scale to highly nonlinear real-world applications. To counteract this, many different techniques have emerged to help ease this problem. Prior work typically has used traditional reinforcement learning strategies such as Deep Q Learning in order to create a desired policy for a robot to interact with its environment ~\cite{rl-for-robots}. However, these techniques routinely suffer from scalability issues due to sparse reward spaces and high data requirements ~\cite{rl-training}. This entails extensive hours of training and simulation, constituting a substantial overhead. Instead, our solution attempts to provide a different architecture that embodies the major principles of reinforcement learning within a slightly different paradigm. 

\begin{figure}[!t]
\centering
\includegraphics[width=\columnwidth]{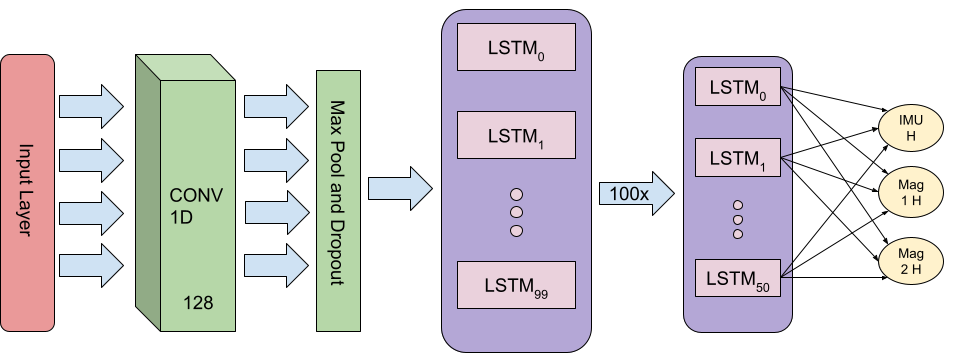}
\vspace{-5mm}
\caption{\textbf{Learning framework. } Sequence-Based Deep Learning Model for Burrowing Depth Approximation. One dimensional convolutional neural network with max pool and dropout layers combined with multiple sequence-based long-short-term memories (LSTMs) used with a final dense layer to approximate depth for a given current motor position.}
\label{ML Schematic}
\end{figure}

Rather than constructing a formal policy array relying on a dense neural network, we adopt a policy representation in the form of a sequence-based model, which incorporates 1D convolutional layers and long-short-term memory (LSTM) attention-based layers. Instead of using a simple dense network, the burrowing robot is trained on a sequence-based network that is designed to approximate the correct sequence of N motor positions required to burrow the snake to a target depth. These neural network architectures have been deployed several times for robotic applications to aid in segmentation and deep vision applications ~\cite{conv-lstm-rob}. This is used to provide a feature extraction pipeline that requires much less data to approximate the value of each motor position in reaching the target depth. Figure \ref{ML Schematic} highlights this model structure. 

The model is trained on multiple trial runs. Each trial run was standardized to last for 60 seconds and categorized into three groups. The first group comprises trial runs in which motor positions are randomly sampled from a normal distribution, denoted as $\theta_i\sim\mathcal{N}(0^\circ, 50^\circ)$. In general, these trials performed the worst of all collected data.

The second group of data came from sinusoidal data. Utilizing sinusoidal data was inspired by biological snakes burrowing in sand and the observation that their movement to achieve vertical self-burrowing was achieved through predictable sinusoidal motion on the horizontal plane \cite{jayne1986kinematics,gidmark2011locomotory}. Thus, for these trials, the motors' positions were classified by a sinusoidal behavior defined as $$\theta_{i,x}=Asin(2\pi ft+i/5f)\ \text{for} \ i=0:4$$ where $A$ is the amplitude, $f$ is the frequency and $t$ is the time. Across all samples of this group, the frequency $f$ was set to 0.2 Hz. In terms of amplitude 

It should be noted that the phase shift to this sin function is implemented in order to ensure a smooth sinusoid across all five motors in the robot. It is also of note that this sinusoidal motion was restricted to the x-axis. For half of these trials, the yaw motors acted alone without the actuation of the pitch motors. For the other half of these trials, pitch motor angles were obtained by sampling $\theta_{i,y}\sim\mathcal{N}(0^\circ, 15^\circ)$ in order to capture the effect of pitch actuation on the self-burrowing process. 

The final and largest group of the data came from using a greedy epsilon approach \cite{greedy-epsilon,liu2022understanding,dos2017adaptive}. This approach commonly used in reinforcement learning seeks to balance the exploration and utilization of the current optimal strategy ~\cite{greedy-epsilon}. A parameter, $\epsilon$, is defined to signify the proportion of time when random data is chosen for the motor actions, commonly referred to as exploratory actions. Conversely, 1- $\epsilon$ is the probability that the ideal sinusoidal waveform is selected \cite{greedy-epsilon}. Thus, for these trials, the motor angles were defined as
\begin{align*}
&\theta_{i,x} =
\begin{cases}
  Asin(2\pi ft+i/5f)\ \text{for} \ i=0:4 &  P(1-\epsilon) \\
  \sim \mathcal{N}(0^\circ, \sigma^2) &  P(\epsilon)
\end{cases} \\
&\theta_{i,y} \sim \mathcal{N}(0^\circ, \sigma^2).
\end{align*}

Across the greedy epsilon trials, the amplitude $A$ was fixed at 30 degrees and the frequency $f$ was fixed at 0.2 Hz. The value of $\epsilon$ was swept from 0.1 to 0.9. This process was repeated twice with low variance, $\sigma^2=15^\circ$, and medium variance, $\sigma^2=15^\circ$, for the randomly sampled variables. We collected 270 trial runs with over 45,000 data points to train our neural network. The sampling rate across all trials averaged out to one sample every 0.39 seconds or 2.53 Hz. 

A unique aspect of burrowing is that the optimal policy may change due to the orientation, position, and displacement of the material the robot is passing through. This is clearly reflected in the generated data where not a single policy from either sinusoidal, greedy-sinusoidal, or random gait patterns is able to reliably burrow to the target depth across all trials.

However, if the sequence-based model is used to optimize the burrowing sequence while also being updated in a reinforcement learning style manner, a consistent policy can be created that allows for correct burrowing to any target depth regardless of the position, forces, or displacement of the packing peanuts and snake.

In order to accomplish this, an optimization strategy is deployed to adapt the sequence-based model to help it both explore and exploit the given policy. This optimization technique is known as a Truncated Newton Conjugate-Gradient (TNC) which dynamically estimates the correct positions of the motors to minimize the distance between the target depth and the current depth. This can be corrected even if the underlying model fails to initially predict a correct position. This optimization process is given below.
\begin{equation}
m^* = \underset{m \in M}{\text{argmin}} \ f(\text{seq}(m_{\text{start}}))
\end{equation}
\vspace{-3mm}
\begin{align*}
\text{s.t.} \ & 0 \leq m \leq 360 \\
\text{s.t.} \ & 0 \leq m_i \leq 360, \ \forall m_i \in m \\
m_{\text{start}} & = p_{\text{prev}} + \text{perturb} \times \text{perturb\_scale}
\end{align*}

\begin{align*}
\hline
&\textbf{Algorithm 1:} \text{ TNC argument minimization}\\
\hline
&\text{Initialize} \ m_0 \\
&\text{For} \ k = 0, 1, 2, \ldots \\
&\quad \nabla f(m_k) \ \text{Compute Gradient} \\
&\quad H(m_k) \ \text{Compute or Approximate Hessian} \\
&\quad d_k = -H(m_k)^{-1} \nabla f(m_k) \ \text{Determine Direction} \\
&\quad \alpha_k \ \text{Determine Optimal Step Size Using Line Search} \\
&\quad m_{k+1} = m_k + \alpha_k d_k \ \text{Update the Guess} \\
&\text{End For}\\
\hline
\end{align*}
\label{TNC argument minimization formula}

In more common terms $m^*$ is the set of motor positions (130 positions for a 60-second experiment), and $p_{\text{prev}}$ is the previous output of the sequence model. $f$ is the prediction function of the model for a given sequence of motor positions $m^*$. Then $\text{perturb}$ is a random perturbation drawn from [-0.5 to 0.5]. Finally, $\text{seq}(m_{\text{start}})$ is a function that takes a motor position m and appends it to the previous positions and returns the resultant sequence. 

The specific TNC optimization formula is expressed in equation (1) and is written as an iterative for loop. optimization problem starts with a perturbation of the last motor position to get an initial guess for the next position. After the position is selected the gradient and the truncated Hessian are computed for the given candidate position. Next, the direction is detected by subtracting the gradient from the inverse of the Hessian. Finally, a line search is used to find the optimal stepping size for the next step and the guess is updated accordingly based on the gradient and Hessian from our previous model. 

It is important to note that this perturbation is used to prevent the snake from solely relying on the model and forces it to search for an optimal policy in order to guarantee that burrowing reaches the proper depth. However, the computation of the Hessian is a very data-intensive mathematical operation, and creating a proper burrowing strategy takes an average of 9.58 minutes of onboard computing on Jetson AGX Nano. This can be a severe bottleneck if multiple burrowing requests are given in rapid succession.

\section{Results and Discussion}

\begin{figure*}[!t]
\centering
\includegraphics[width=\textwidth]{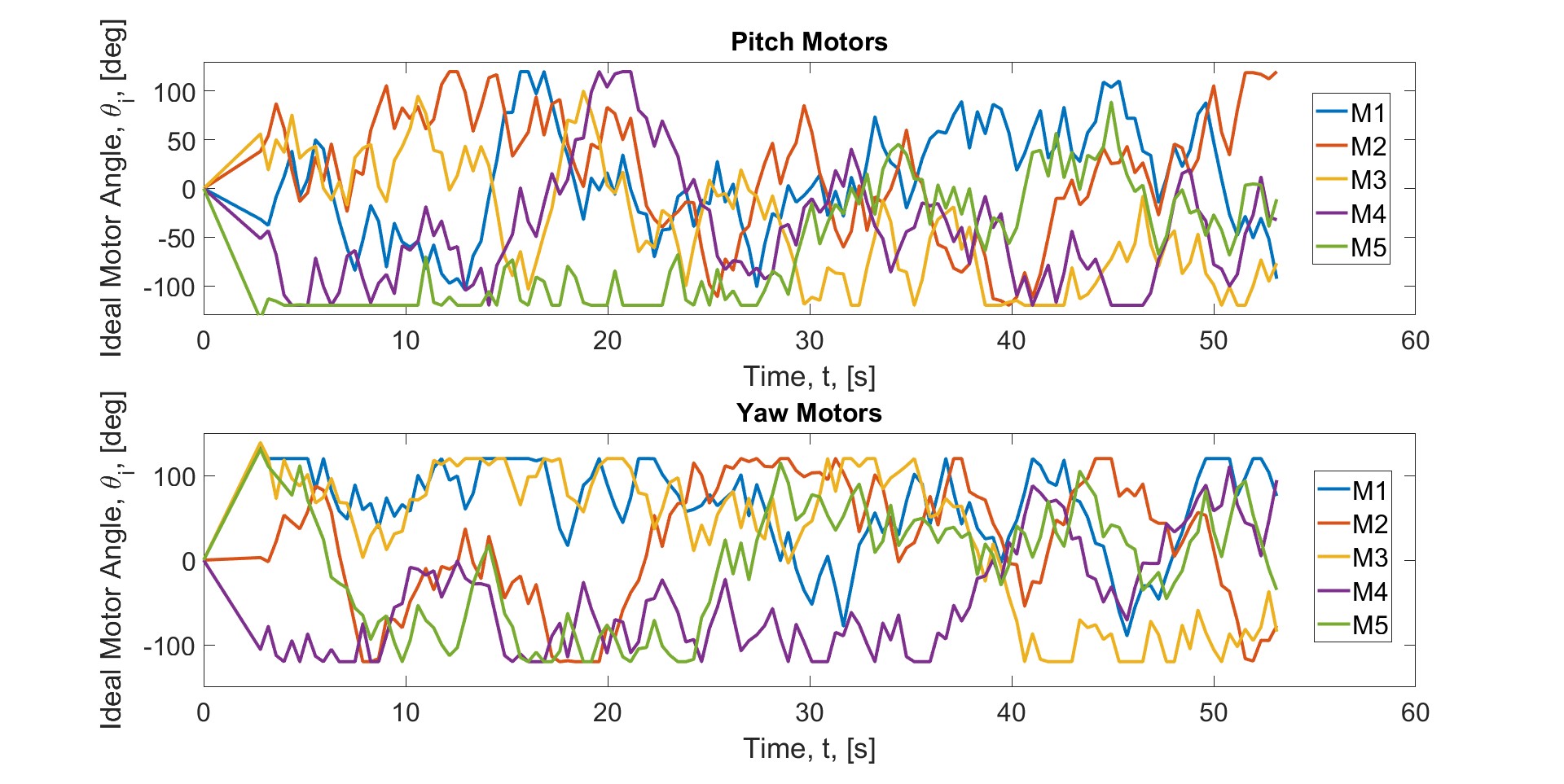}
\caption{Ideal burrowing strategy obtained through the learning process. All motors begin at $0^\circ$ and move to 130 unique positions in order to achieve efficient self-burrowing.}
\label{optimal}
\end{figure*}

After training our model based on the collected data and conducting TNC argument minimization as described in the previous section, we generated a set of 130 angular positions for all ten motors. A visual representation of this burrowing strategy is depicted in Figure \ref{optimal}. In total, this sequence lasted just under a minute to correspond well with the training data. It was observed that extreme motor angles would result in self-collisions of the links of the robot that would cause the robot to stall. This collision did not appear if the motor angle remained between the range of $-120^\circ$ and $120^\circ$. Thus, in training, we penalized motor values outside this range and clipped motor values that exceeded this range.

The optimal gait pattern begins by sending most of the motor positions to extreme motor positions Visually, the effect of this is to greatly shorten the distance from tip to tail. From there, the body wriggles back and forth to lower the robot deeper into the granular medium. In this way, the behavior learned by the robot greatly resembles techniques used by vipers for vertical self-burrowing in sand \cite{Vipers}. The emergence of behavior similar to that observed in biology is further validation of the efficacy of this technique.

For the tests, a target depth of 15 centimeters above the base of the testing environment was chosen. This height was chosen because it was at least five centimeters deeper than any of the methods used in the training data could reliably reach. The closer the robot moved to the base of the testing environment, the more significant boundary effects become making burrowing more challenging \cite{boundary}.

As previously mentioned, the test data failed to reliably adapt to the unique burrowing requirements in each trial run which required the robotic snake to get to a target depth of 15 cm from the bottom of the bin. In isolated incidents the snake occasionally was able to reach the target depth; however, it was not able to do this in every trial, and, when it did, it often moved up away from this target depth. 

\begin{figure}[t]
\centering
\includegraphics[width=\columnwidth]{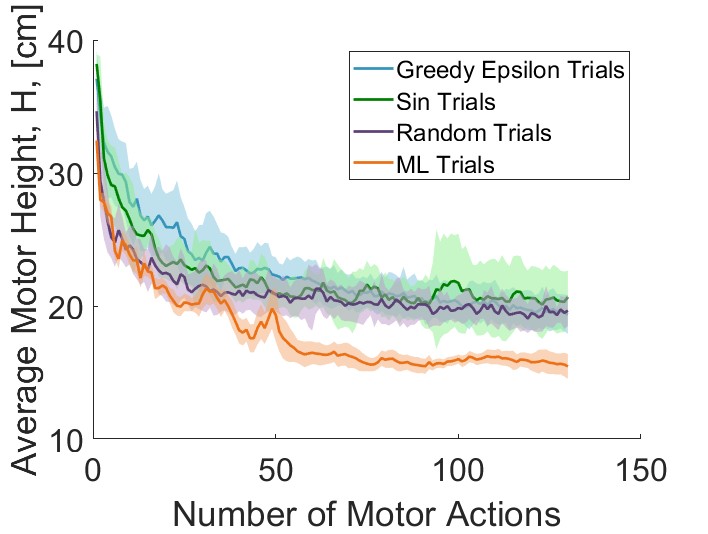}
\caption{The plot above demonstrates the effectiveness of vertical self-burrowing for each class of training data as the strategy obtained through Machine Learning. For each class of data, the error band shows one standard deviation from the mean as a function of the number of motor actions.}
\label{Statistical Results}
\end{figure}

The optimized model combined with the TNC optimization technique was able to reach a targeted depth of 15 cm in every trial it performed. This allowed for precise and controlled burrowing by using a sequence-based model combined with an optimization technique that dynamically minimizes the current position subtracted from the target position. A comprehensive comparison between the training data and the strategy achieved through our ML process can be observed in Figure \ref{Statistical Results}.

Three selected experiments each with ten trials are shown in Figure \ref{Statistical Results}. As evidenced from the plot, the ML trials were successful in reaching the target depth of 15 cm from the bottom of the bin while the greedy epsilon, sinusoidal trials, and random trials were all unsuccessful.  In other words, the burrowing strategy obtained from machine learning is able to burrow deeper and more efficiently than any other strategy we employed. Secondly, it is important to notice that our solution has significantly lower variance demonstrating that the policy was able to reliably adapt unique burrowing attempts to successfully reproduce improved vertical self-burrowing. 

Although our techniques are successful there are still many open research directions. We have identified three general directions: 

\begin{enumerate}
    \item \textbf{Power aware model compression and quantization}
    \item \textbf{NVIDIA-based Isaac Simulation for synthetic data generation}
    \item \textbf{Replacing LSTMs with Transformer Based Attention Mechanism}
\end{enumerate}

First and foremost, ML models have very high computation requirements, particularly with attention-based models. Edge robotics must be aware of this computational expense and provide accurate solutions within low-power envelopes. This is achieved through model compression and quantization to reduce power consumption as much as possible while keeping a minimum accuracy. Secondly, our robot does have certain drawbacks in regard to its data acquisition pipeline. Building robots and then deploying them is not a scalable process since the cost to develop a robot is paid without a strong understanding of how well the robot will perform on a task. This can lead to capital losses and inefficiencies in robot design. To combat this we have created virtual environments of our robots in NVIDIA Omniverse and are using the NVIDIA Isaac Gym to train the robot in a simulated environment which will be deployed to a real-world robotic environment where results between simulation and the real-world environment can be compared. Our final future research direction will explore the usage of transformers for burrowing depth estimation. BERT and several other attention-based models that are more advanced than LSTMs were utilized with limited efficacy. This is primarily due to the fact that transformers require much more data despite their ability to generalize more effectively. We are generating more data to test Transformers for our application.

\section{Conclusion}
The promising outcomes achieved in vertical burrowing serve as motivation to extend the training methodology for the robot to perform horizontal burrowing in future research. These advancements contribute to the field of robotics and have implications for applications requiring efficient and adaptive burrowing capabilities in challenging environments.

\begin{figure}[!t]
\centering
\includegraphics[width=\columnwidth]{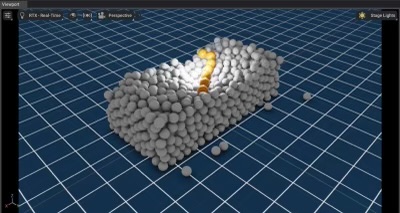}
\caption{  Our experimental environment is modeled in Nvidia Omniverse with granular media modeled by white spheres and the snake robot modeled in yellow. We will leverage this platform to augment our training dataset by executing simulated trial runs in order to refine a more optimal training methodology.}
\label{sim}
\end{figure}

While the results in this paper show significant improvement from random actions, we recognize that better results require the acquisition of more training data. With this in mind, we have begun to simulate the testing environment used in this paper in the Nvidia Omniverse simulation environment as seen in Fig. \ref{sim}. This will allow us to generate a much larger volume of data to develop even better strategies in simulation, and then validate those results in real-world experiments. Furthermore, we hope to eventually move to finer media to more closely approximately burrowing in natural environments such as dirt and sand. 

Finally, in the future, we will explore efforts to minimize the energy needed for neural networks to function without greatly impacting performance. Because of the computing power and energy consumed to run neural networks, it is difficult to deploy them in mobile robots. However, these models can be greatly simplified through quantization. Reducing energy consumption and increasing the efficiency of neural makes them more deployable in mobile robots. 

\bibliographystyle{IEEEtran}

\begin{thebibliography}{10}
\providecommand{\url}[1]{#1}
\csname url@samestyle\endcsname
\providecommand{\newblock}{\relax}
\providecommand{\bibinfo}[2]{#2}
\providecommand{\BIBentrySTDinterwordspacing}{\spaceskip=0pt\relax}
\providecommand{\BIBentryALTinterwordstretchfactor}{4}
\providecommand{\BIBentryALTinterwordspacing}{\spaceskip=\fontdimen2\font plus
\BIBentryALTinterwordstretchfactor\fontdimen3\font minus \fontdimen4\font\relax}
\providecommand{\BIBforeignlanguage}[2]{{%
\expandafter\ifx\csname l@#1\endcsname\relax
\typeout{** WARNING: IEEEtran.bst: No hyphenation pattern has been}%
\typeout{** loaded for the language `#1'. Using the pattern for}%
\typeout{** the default language instead.}%
\else
\language=\csname l@#1\endcsname
\fi
#2}}
\providecommand{\BIBdecl}{\relax}
\BIBdecl

\bibitem{li2013terradynamics}
C.~Li, T.~Zhang, and D.~I. Goldman, ``A terradynamics of legged locomotion on granular media,'' \emph{science}, vol. 339, no. 6126, pp. 1408--1412, 2013.

\bibitem{pradhan2019mass}
S.~Pradhan and T.~Siddique, ``Mass wasting: an overview,'' \emph{Landslides: Theory, Practice and Modelling}, pp. 3--20, 2019.

\bibitem{maladen2011mechanical}
R.~D. Maladen, Y.~Ding, P.~B. Umbanhowar, A.~Kamor, and D.~I. Goldman, ``Mechanical models of sandfish locomotion reveal principles of high performance subsurface sand-swimming,'' \emph{Journal of The Royal Society Interface}, vol.~8, no.~62, pp. 1332--1345, 2011.

\bibitem{martinez2022bio}
A.~Martinez, J.~DeJong, I.~Akin, A.~Aleali, C.~Arson, J.~Atkinson, P.~Bandini, T.~Baser, R.~Borela, R.~Boulanger \emph{et~al.}, ``Bio-inspired geotechnical engineering: Principles, current work, opportunities and challenges,'' \emph{G{\'e}otechnique}, vol.~72, no.~8, pp. 687--705, 2022.

\bibitem{roots}
I.~Taylor, K.~Lehner, E.~McCaskey, N.~Nirmal, Y.~Ozkan-Aydin, M.~Murray-Cooper, R.~Jain, E.~W. Hawkes, P.~C. Ronald, D.~I. Goldman, and P.~N. Benfey, ``Mechanism and function of root circumnutation,'' \emph{Proceedings of the National Academy of Sciences}, vol. 118, no.~8, p. e2018940118, 2021.

\bibitem{Naclerio}
N.~D. Naclerio, A.~Karsai, M.~Murray-Cooper, Y.~Ozkan-Aydin, E.~Aydin, D.~I. Goldman, and E.~W. Hawkes, ``Controlling subterranean forces enables a fast, steerable, burrowing soft robot,'' \emph{Science Robotics}, vol.~6, no.~55, p. eabe2922, 2021.

\bibitem{omori2008locomotion}
H.~Omori, T.~Hayakawa, and T.~Nakamura, ``Locomotion and turning patterns of a peristaltic crawling earthworm robot composed of flexible units,'' in \emph{2008 IEEE/RSJ International Conference on Intelligent Robots and Systems}.\hskip 1em plus 0.5em minus 0.4em\relax IEEE, 2008, pp. 1630--1635.

\bibitem{Tao_2021}
J.~J. Tao, ``Burrowing soft robots break new ground,'' \emph{Science Robotics}, vol.~6, no.~55, 2021.

\bibitem{Liu}
B.~Liu, Y.~Ozkan-Aydin, D.~I. Goldman, and F.~L. Hammond, ``Kirigami skin improves soft earthworm robot anchoring and locomotion under cohesive soil,'' in \emph{2019 2nd IEEE International Conference on Soft Robotics (RoboSoft)}, 2019, pp. 828--833.

\bibitem{Russell}
\BIBentryALTinterwordspacing
R.~A. Russell, ``Crabot: A biomimetic burrowing robot designed for underground chemical source location,'' \emph{Advanced Robotics}, vol.~25, no. 1-2, pp. 119--134, 2011. [Online]. Available: \url{https://doi.org/10.1163/016918610X538516}
\BIBentrySTDinterwordspacing

\bibitem{Treers}
\BIBentryALTinterwordspacing
L.~K. Treers, B.~McInroe, R.~J. Full, and H.~S. Stuart, ``Mole crab-inspired vertical self-burrowing,'' \emph{Frontiers in Robotics and AI}, vol.~9, 2022. [Online]. Available: \url{https://www.frontiersin.org/articles/10.3389/frobt.2022.999392}
\BIBentrySTDinterwordspacing

\bibitem{Bagheri2023}
\BIBentryALTinterwordspacing
H.~Bagheri, D.~Stockwell, B.~Bethke, N.~K. Okwae, D.~Aukes, J.~Tao, and H.~Marvi, ``A bio-inspired helically driven self-burrowing robot,'' \emph{Acta Geotechnica}, Apr 2023. [Online]. Available: \url{https://doi.org/10.1007/s11440-023-01882-9}
\BIBentrySTDinterwordspacing

\bibitem{PhysRevE.105.064902}
\BIBentryALTinterwordspacing
K.~Lee and R.~C. Hurley, ``Force inference in granular materials: Uncertainty analysis and application to three-dimensional experiment design,'' \emph{Phys. Rev. E}, vol. 105, p. 064902, Jun 2022. [Online]. Available: \url{https://link.aps.org/doi/10.1103/PhysRevE.105.064902}
\BIBentrySTDinterwordspacing

\bibitem{rl-for-robots}
J.~Kober, J.~A. Bagnell, and J.~Peters, ``Reinforcement learning in robotics: A survey,'' \emph{The International Journal of Robotics Research}, vol.~32, no.~11, pp. 1238--1274, 2013.

\bibitem{rl-training}
J.~Ibarz, J.~Tan, C.~Finn, M.~Kalakrishnan, P.~Pastor, and S.~Levine, ``How to train your robot with deep reinforcement learning: lessons we have learned,'' \emph{The International Journal of Robotics Research}, vol.~40, no. 4-5, pp. 698--721, 2021.

\bibitem{conv-lstm-rob}
A.~Cura, H.~K{\"u}{\c{c}}{\"u}k, E.~Ergen, and {\.I}.~B. {\"O}ks{\"u}zo{\u{g}}lu, ``Driver profiling using long short term memory (lstm) and convolutional neural network (cnn) methods,'' \emph{IEEE Transactions on Intelligent Transportation Systems}, vol.~22, no.~10, pp. 6572--6582, 2020.

\bibitem{jayne1986kinematics}
B.~C. Jayne, ``Kinematics of terrestrial snake locomotion,'' \emph{Copeia}, pp. 915--927, 1986.

\bibitem{gidmark2011locomotory}
N.~J. Gidmark, J.~A. Strother, J.~M. Horton, A.~P. Summers, and E.~L. Brainerd, ``Locomotory transition from water to sand and its effects on undulatory kinematics in sand lances (ammodytidae),'' \emph{Journal of Experimental Biology}, vol. 214, no.~4, pp. 657--664, 2011.

\bibitem{greedy-epsilon}
M.~Tokic, ``Adaptive $\varepsilon$-greedy exploration in reinforcement learning based on value differences,'' in \emph{Annual Conference on Artificial Intelligence}.\hskip 1em plus 0.5em minus 0.4em\relax Springer, 2010, pp. 203--210.

\bibitem{liu2022understanding}
F.~Liu, L.~Viano, and V.~Cevher, ``Understanding deep neural function approximation in reinforcement learning via $\epsilon$-greedy exploration,'' \emph{Advances in Neural Information Processing Systems}, vol.~35, pp. 5093--5108, 2022.

\bibitem{dos2017adaptive}
A.~dos Santos~Mignon and R.~L. d.~A. da~Rocha, ``An adaptive implementation of $\varepsilon$-greedy in reinforcement learning,'' \emph{Procedia Computer Science}, vol. 109, pp. 1146--1151, 2017.

\bibitem{Vipers}
B.~Young, M.~Morain, and R.~Wood, ``Vertical burrowing in the saharan sand vipers (cerastes),'' \emph{Copeia}, vol. 2003, pp. 131--137, 02 2003.

\bibitem{boundary}
\BIBentryALTinterwordspacing
G.~Marketos and M.~D. Bolton, ``Flat boundaries and their effect on sand testing,'' \emph{International Journal for Numerical and Analytical Methods in Geomechanics}, vol.~34, no.~8, pp. 821--837, 2010. [Online]. Available: \url{https://onlinelibrary.wiley.com/doi/abs/10.1002/nag.835}
\BIBentrySTDinterwordspacing

\end{thebibliography}

\end{document}